# Fully Autonomous Z-Score–Based TinyML Anomaly Detection on Resource-Constrained MCUs Using Power Side-Channel Data

Abdulrahman Albaiz
Department of Computer Science
& Engineering
Wright State University
Dayton, Ohio, USA
Email: albaiz.2@wright.edu

Fathi Amsaad
Department of Computer Science
& Engineering
Wright State University
Dayton, Ohio, USA
Email: fathi.amsaad@wright.edu

*Abstract—* **This paper presents a fully autonomous Tiny Machine Learning (TinyML) Z-Score–based anomaly detection system deployed on a low-power microcontroller for real-time monitoring of appliance behavior using power side-channel data. Unlike existing Internet of Things (IoT) anomaly detection approaches that rely on offline training or cloud-assisted analytics, the proposed system performs both model training and inference directly on a resource-constrained microcontroller without external computation or connectivity. The system continuously samples current consumption, computes Root Mean Square (RMS) values on-device, and derives statistical parameters during an initial training phase. Anomalies are detected using lightweight Z-Score thresholds, enabling interpretable and computationally efficient inference suitable for embedded deployment. The architecture was implemented on an STM32-based platform and evaluated using a 14-day dataset collected from a household mini-fridge under normal operation and controlled anomaly conditions. Results demonstrate perfect detection performance, with Precision and Recall of 1.00, inference latencies on the order of tens of microseconds, and a total memory footprint of approximately 3.3 KB SRAM and 63 KB Flash. These results confirm that robust and fully autonomous TinyML anomaly detection can be achieved on low-cost microcontrollers. Future work includes extending the framework to incorporate additional lightweight models and multi-device learning scenarios.**

*Keywords— TinyML, Anomaly Detection, Microcontrollers, Edge Computing, Z-Score, Power Monitoring, RMS Current, Side-Channel Sensing*

## I. INTRODUCTION

TinyML has enabled machine learning to operate directly on resource-constrained microcontrollers (MCUs), allowing edge devices to execute inference without reliance on cloud connectivity [1]–[4]. This capability is critical for applications requiring low latency, high reliability, and strong data privacy, where cloud-based architectures introduce communication delays, cost, and vulnerability to network disruptions.

Among various sensing modalities, power side-channel measurements—specifically electrical current consumption—offer a lightweight and non-intrusive method [5], [6] for characterizing the operational behavior of appliances and embedded systems. Changes in current flow reflect internal mechanical and electrical states, enabling the detection of abnormal behavior without the need for additional sensors. This makes current sensing an attractive candidate for edge-based anomaly detection in resource-limited environments.

Despite the potential advantages, existing Internet of Things (IoT) anomaly detection solutions frequently rely on offline training or cloud-assisted analytics [4], limiting their autonomy and effectiveness in real-world deployments. Complex ML models used in prior work often exceed the computational budget of low-power MCUs [7]–[10] or require external infrastructure for model training. These constraints highlight the need for a fully autonomous anomaly detection method that operates entirely on the microcontroller.

To address this gap, this paper presents a fully autonomous Z-Score–based anomaly detection system that performs both training and inference directly on an MCU. Using Root Mean Square (RMS) current measurements captured from a low-cost ACS712 sensor, the system learns the normal operating profile of an appliance and subsequently identifies anomalies without external supervision or cloud connectivity. This design enables reliable long-term monitoring suitable for embedded field deployments.

Contributions:

In this work, we provide the following contributions

1. A lightweight Z-Score anomaly detection model optimized for TinyML-class MCUs.

2. An autonomous workflow enabling the MCU to compute statistical parameters during an initial training phase and transition seamlessly into inference mode.

3. A complete end-to-end implementation including sensing, RMS computation, timestamping, and MicroSD-based data logging.

4. A real-world evaluation using a 14-day dataset collected from a household appliance under normal and controlled anomalous conditions.

5. Detailed runtime and memory profiling demonstrating the feasibility of MCU-only anomaly detection.

The primary contribution of this work lies in the autonomous TinyML system design and on-device learning pipeline, rather than proposing a new anomaly detection algorithm.

## II. RELATED WORK

### A. TinyML-Based Edge Anomaly Detection

TinyML research has investigated running machine learning models on MCUs for tasks such as vibration monitoring, environmental sensing, and acoustic analysis. Prior approaches frequently employ clustering, autoencoders, or neural models; however, these methods often require offline training, cloud inference, or memory footprints unsuitable for low-power MCUs. Existing implementations do not provide a self-contained anomaly detection pipeline [7]–[10] that performs both model training and inference directly on the device. Broader surveys on anomaly detection in smart environments and IoT systems further highlight the need for robust and efficient detection methods on constrained devices [4].

### B. Power Side-Channel Sensing

Power side-channel analysis has been widely used [6] in hardware security and system diagnostics, where current traces reveal operational state transitions or faulty behavior. Studies demonstrated the effectiveness of current signatures [5], [6] for identifying anomalous patterns. However, these techniques typically rely on high-resolution measurement equipment or offline analysis, limiting their applicability to embedded IoT deployments.

### C. Statistical Models on MCUs

Statistical anomaly detection models, such as Z-Score [11], [12], are well suited for embedded platforms due to their simplicity and minimal computational overhead. While statistical methods have been applied in IoT anomaly detection, prior work generally performs computations off-device or uses MCUs solely for data acquisition. A fully autonomous Z-Score implementation operating entirely on an MCU has not been reported in the literature.

### D. Gap in Current Research

Existing TinyML anomaly detection systems demonstrate that training and inference can be executed directly on microcontrollers for industrial and IoT deployments, often using tree ensembles, neural networks, or online learning mechanisms [7]–[10]. Similarly, power side-channel signals such as current traces have been used to detect abnormal behavior in hardware and embedded platforms [5], [6]. However, to the best of our knowledge, no prior work has reported a fully documented Z-Score–based anomaly detector that (i) learns statistical parameters entirely on-device, (ii) operates continuously on a low-cost MCU, and (iii) detects anomalous operating behavior from RMS current features without relying on external computation. This work addresses this gap. Unlike prior MCU-based anomaly detectors, this work introduces a multi-feature, five-dimensional Z-score model that computes a composite anomaly score entirely on-device.

## III. SYSTEM ARCHITECTURE

The system performs continuous monitoring of appliance current consumption and processes the data entirely on the MCU to identify abnormal behavior. The architecture consists of a sensing front-end, embedded computation pipeline, and local storage components. The overall hardware–software architecture of the system is summarized in Fig. 1.

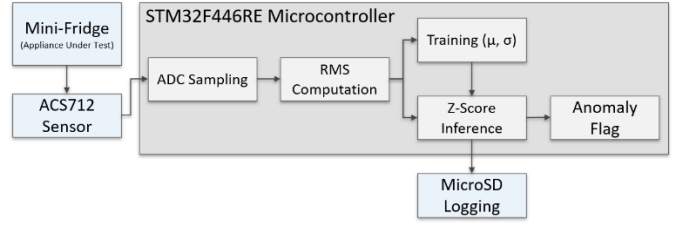

Fig. 1. System architecture diagram.

### A. Hardware Components

The hardware configuration includes:

- STM32 NUCLEO-F446RE microcontroller (ARM Cortex-M4).
- ACS712 Hall-effect current sensor for non-intrusive AC current measurement.
- MicroSD storage module for persistent logging.
- A simple resistor-based noise-reduction network stabilizing the analog sensor output.

This minimal hardware configuration supports low-power, continuous monitoring while maintaining low system complexity [13]–[15].

### B. Data Acquisition Workflow

The ACS712 sensor output is sampled by the STM32 analog-to-digital converter (ADC) at high frequency. Every 1000 samples are aggregated into a single RMS current measurement, which characterizes the appliance's operational state. A real-time clock provides timestamps, and each RMS entry is stored locally on the MicroSD card for future offline analysis or performance auditing.

### C. RMS Computation

The RMS current is computed as shown in (1):

$$\text{RMS} = \sqrt{\tfrac{1}{N}\sum x_i^2} \qquad (1)$$

Empirical evaluation showed that RMS values below 0.06–0.09 A correspond to compressor OFF states, while values around 0.86–0.88 A indicate ON cycles. These RMS measurements serve as the inputs to the Z-Score detector.

### D. Autonomous Operation Workflow

The system operates in two autonomous phases:

*1) Training Phase:* The MCU collects RMS data during an initial operating period and computes the mean and standard deviation representing normal behavior.

*2) Inference Phase:* After completing the training period, the MCU transitions into inference mode and computes a Z-Score for each completed compressor ON cycle using its duration, flagging anomalies when values exceed a predefined threshold.

This workflow enables fully on-device anomaly detection without reliance on cloud infrastructure or external supervisory control. In addition to Z-Score–based detection, the system implements a lightweight watchdog rule that flags a power-off anomaly whenever the compressor remains in the OFF state for an excessively long period (approximately 60 minutes).

## IV. Z-Score Model Design

The anomaly detection model is based on the standard Z-Score statistical measure, which quantifies how far a new observation deviates from the learned mean of normal behavior. This model is well suited for microcontrollers due to its simplicity, low computational overhead, and constant-time inference.

On the MCU, each compressor cycle is summarized by a compact five-dimensional feature vector that captures both instantaneous and temporal behavior: RMS current, mean RMS over the window, RMS standard deviation, RMS slope, and the ON duration of the compressor cycle (Duration_On). During detection, a per-feature Z-Score is computed using the learned mean and standard deviation for each dimension, and a composite anomaly score is obtained as the average of the absolute Z-Scores across all five features, assigning equal weight to each feature. This design allows the model to exploit multiple characteristics of the power trace while keeping computation and memory overhead small.

RMS samples are first classified into ON/OFF compressor states. For each ON cycle, its duration is computed and incorporated as one of the feature dimensions within the Z-Score model. This converts temporal compressor behavior into a statistical anomaly detection problem.

For visualization purposes, the Z-score trend shown in Fig. 4 illustrates the ON-duration feature only; this one-dimensional marginal closely reflects the behavior of the full composite 5-feature anomaly score used in deployment.

### A. Training Phase

During an initial training period, the MCU processes feature records produced by the data pipeline. For each compressor cycle, it extracts a five-dimensional feature vector consisting of RMS current, mean RMS, RMS standard deviation, RMS slope, and compressor ON duration (Duration_On). For each feature dimension, the MCU incrementally accumulates the required statistics to support online computation. At the end of training, the mean μ and standard deviation σ for each feature are computed using the standard formulas in (2) and (3), respectively. These statistics define the normal operating distribution used during inference.

$$\mu = \frac{1}{M}\sum_{i=1}^{M} x_i \quad (2)$$

$$\sigma = \sqrt{\frac{1}{M}\sum_{i=1}^{M}(x_i - \mu)^2} \quad (3)$$

To reduce memory usage and avoid storing the full training dataset, the MCU computes these parameters incrementally. This enables the training process to run efficiently even with limited static random-access memory (SRAM).

### B. Inference Phase

Once the training phase is complete, the MCU transitions automatically into inference mode. For each new feature record, the same five-dimensional feature vector is recomputed and a Z-Score is evaluated independently for each feature using (4). The absolute Z-Scores are then averaged across the five dimensions to produce a composite anomaly score. If this composite score exceeds a fixed threshold (2.5 in our experiments), the corresponding samples are treated as anomalous and an anomaly streak counter is updated. When analyzing the one-dimensional Z-Score distribution shown in Fig. 4, $x$ corresponds to the compressor ON duration; in the deployed composite model, $x_i$ is defined separately for each feature dimension.

$$z = \frac{x - \mu}{\sigma} \quad (4)$$

Equal weighting was chosen to minimize model complexity and avoid feature-specific tuning on the MCU; investigating adaptive or learned weighting schemes is left for future work. The anomaly threshold (2.5) was selected empirically based on observed separation between normal and anomalous compressor cycles during training; a formal sensitivity or ROC analysis is left for future work.

### C. Advantages of the Z-Score Model on MCUs

The Z-Score method offers several advantages for embedded anomaly detection, including minimal memory usage (only mean and standard deviation per feature), deterministic real-time execution, transparent interpretability, and robustness to minor power fluctuations. These properties make it well suited for low-power MCUs deployed in unattended environments.

## V. Implementation on STM32 MCU

The anomaly detection pipeline is implemented entirely on the STM32 NUCLEO-F446RE microcontroller. The system executes data acquisition, RMS computation, Z-Score calculation, and event logging without reliance on external processing.

### A. ADC Configuration and Sampling

The ACS712 analog output is sampled using the STM32's 12-bit ADC at a high sampling frequency sufficient to capture the 60 Hz AC waveform envelope [14]. ADC readings are collected into a buffer of 1000 samples per RMS computation cycle. The sampling routine is interrupt-driven to maintain timing accuracy and minimize CPU overhead.

### B. RMS Computation Pipeline

For each 1000-sample block, the MCU computes:

1. Square of each sample
2. Mean of squared values
3. Square root of the mean

This produces a single RMS value representing the appliance's instantaneous power consumption. The MCU computes RMS values continuously throughout training and inference.

*C. Timestamping and Data Logging*

A real-time clock (RTC) provides timestamps for each RMS value. The MCU logs entries to a MicroSD card in CSV format, enabling long-term storage for evaluation and analysis. Each record includes the timestamp, RMS current value, Z-Score, and anomaly flag. The MicroSD module is used solely for logging and is not required for real-time anomaly detection. The system operates fully on the MCU even without external storage.

*D. Memory Optimization*

The implementation is optimized to fit the constraints of the STM32:

- Only a fixed-size ADC buffer is maintained in SRAM
- Intermediate values are computed on-the-fly without storing sample histories
- The training phase uses incremental mean and variance updates, requiring no arrays
- All variables are stored as 32-bit floating-point or integer values depending on precision requirements

The complete program, including data logging, training logic, RMS computation, and inference, fits within the available Flash and SRAM resources of the MCU.

*E. Runtime Profiling*

Execution timing was evaluated to ensure the system meets real-time constraints. Z-Score inference executes well within the available processing window, providing ample CPU headroom for continuous operation. RMS computation and MicroSD logging introduce negligible computational load relative to the 30-second sampling interval, confirming that the system can operate continuously without timing violations.

## VI. EXPERIMENTAL SETUP

*A. Data Collection Environment*

The anomaly detection system was deployed on a mini-fridge over a continuous 14-day period. RMS current values were computed from 1000-sample ADC windows and logged once every 30 seconds. This dataset includes both normal compressor cycles and controlled anomalous conditions used to evaluate detection performance.

A representative RMS current profile collected during the deployment is shown in Fig. 2, illustrating normal and controlled abnormal compressor behaviors.

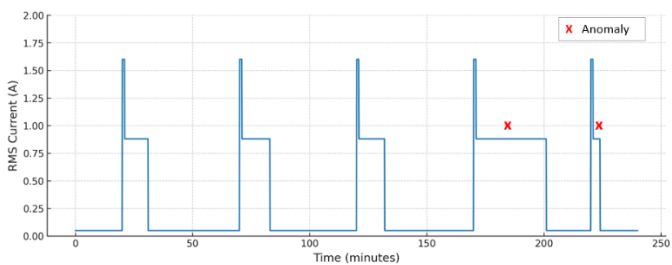

Fig. 2. Example RMS current profile.

All computations were executed directly on the STM32 microcontroller. The entire anomaly detection pipeline—including RMS computation, statistical parameter learning, Z-Score inference, and anomaly flagging—was implemented in C and executed on the MCU without any offloading to a PC or cloud system. This ensures that all reported results reflect true on-device performance and demonstrate the feasibility of fully autonomous anomaly detection on resource-constrained hardware.

The physical experimental setup used during the 14-day deployment is shown in Fig. 3.

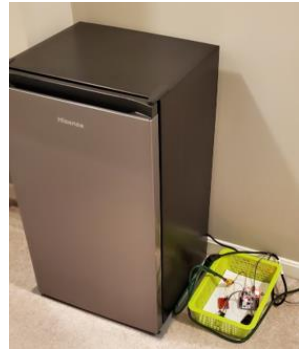

Fig. 3. Experimental setup used in the 14-day deployment.

*B. Normal Operation Profile*

During normal operation, the compressor alternated between two states:

- On-state: RMS ≈ 0.86 – 0.88 A
- Off-state: RMS ≈ 0.07 A

These patterns formed the baseline for training the Z-Score model.

*C. Induced Anomalous Scenarios*

Several controlled abnormal conditions were introduced during the 14-day deployment to evaluate detection performance:

- Increasing the thermostat setting to force extended compressor runtime
- Opening the fridge door to induce cooling inefficiency
- Power disruption events via thermostat shutdown

Each anomaly produced distinctive RMS deviations from the learned baseline.

*D. Evaluation Metrics*

The following metrics are used to quantify anomaly detection performance:

- *Precision:* the proportion of detected anomalies that were true anomalies (values range from 0 to 1).
- *Recall:* the proportion of real anomalies that were successfully detected (values range from 0 to 1).
- *F1 score:* the harmonic mean of precision and recall.

- *Detection delay:* the time elapsed between the onset of an anomalous condition and when the system flags it as an anomaly.

These metrics quantify both reliability and responsiveness of the system.

### E. Dataset Summary

The complete dataset includes:

- 14 days of continuous RMS measurements
- More than 40,000 RMS samples
- Multiple induced anomalous windows
- Full timestamped logs for offline analysis

This real-world dataset provides a robust basis for evaluating the proposed fully autonomous anomaly detection system.

## VII. RESULTS

This section presents the performance evaluation of the fully autonomous Z-Score anomaly detection system deployed on the STM32 microcontroller. Results include anomaly detection accuracy across induced scenarios, runtime performance, and memory usage.

### A. Detection of Anomalous Events

The system successfully detected all induced abnormal operating conditions during the 14-day deployment. Each anomaly produced changes in compressor ON duration that resulted in Z-Score values exceeding the anomaly threshold. Key detection outcomes are summarized below:

- *Thermostat Adjustment:* Increasing the thermostat setting resulted in a prolonged compressor runtime lasting approximately 5 hours. The system detected the anomaly as soon as the prolonged ON duration exceeded the learned statistical baseline.

- *Door-Open Events:* Two door-open events were introduced, each lasting approximately 15 minutes. Both resulted in extended compressor runtime and delayed transitions to the off-state. The Z-Score model correctly identified these deviations as anomalies.

- *Power Disruptions:* Temporary loss of power generated abrupt RMS drops followed by irregular recovery patterns. The system detected these events via the watchdog mechanism, which triggers an anomaly when the compressor remains OFF for longer than the allowed threshold (~60 minutes).

Across all scenarios, the model demonstrated consistent and reliable performance, detecting every introduced anomaly without false positives during stable compressor cycles.

### B. Quantitative Performance Metrics

Across the induced anomaly scenarios evaluated during the 14-day deployment, the system achieved event-level Precision = 1.00, Recall = 1.00, and F1 = 1.0. These results indicate correct detection of all manually induced anomalies with no false alarms during normal operation. However, the evaluation is limited to a single appliance and controlled fault scenarios, and the reported metrics should be interpreted as proof-of-concept performance rather than statistically generalizable accuracy.

### C. Inference Latency

The Z-score model was evaluated using compressor ON durations extracted from the RMS current trace. Normal cycles produce low composite Z-scores, while the controlled long-ON and short-ON scenarios yield statistically significant deviations that exceed the detection threshold. Fig. 4 shows the Z-score distribution computed over compressor ON durations, where the anomalous events cross the ±2.5 threshold and are correctly identified as anomalies. The ON-duration feature is shown for clarity, as it strongly correlates with the composite five-dimensional anomaly score used during deployment.

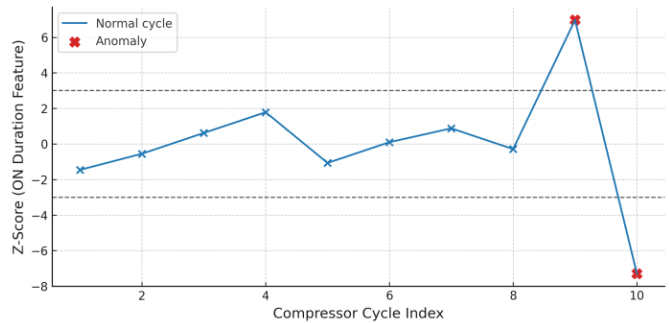

Fig. 4. Z-Score of the ON-duration feature (1D marginal of the 5-feature model).

Inference latency was measured on the STM32 by timing the Z-Score computation and anomaly thresholding per RMS value. Results show:

- Z-Score inference time: ~20–40 µs
- RMS computation time for 1000 samples: Within real-time constraints and completed well before the next sampling window
- MicroSD logging overhead: Did not interfere with the real-time anomaly detection loop

These measurements confirm that the on-device pipeline is capable of continuous operation without risking missed samples or timing delays.

### D. Memory and Flash Footprint

The runtime and memory characteristics of the implementation are summarized in Fig. 5.

The final compiled firmware utilized ~63 KB of Flash memory and ~3.3 KB of SRAM. The reported SRAM usage includes the complete 1000-sample ADC buffer, variables used for RMS computation, Z-Score model parameters, temporary computation structures, and system stack memory. These values demonstrate that the fully autonomous anomaly detection system fits comfortably within the resource limits of low-cost microcontrollers. Compared to other TinyML deployments that rely on tree ensembles or neural models requiring tens of kilobytes of RAM [7], [10], the proposed implementation achieves equivalent functionality with substantially lower memory requirements.

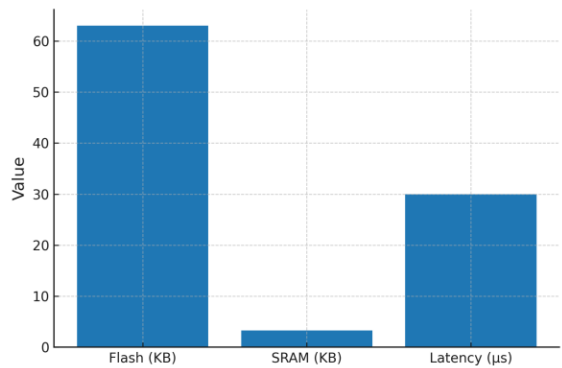

Fig. 5. MCU runtime and memory usage.

## VIII. DISCUSSION

The results demonstrate that the proposed Z-Score–based detector provides a reliable, interpretable, and computationally efficient solution for identifying abnormal power consumption behavior on resource-constrained embedded hardware. Across all induced anomaly scenarios, the system achieved consistent detection performance with no false positives observed during normal compressor operation. The OFF-state watchdog is implemented as a complementary rule-based safeguard and is reported separately from the Z-Score detector, reflecting practical embedded system design considerations rather than statistical anomaly modeling.

A key strength of the proposed approach is its complete autonomy. All stages of data processing, feature extraction, statistical learning, inference, and event logging are executed locally on the MCU without reliance on cloud connectivity or offline computation. This enables long-term deployment in environments where network access is unavailable or unreliable, while maintaining deterministic real-time performance and low resource usage.

The multi-feature Z-Score model learns only ten statistical parameters (mean and standard deviation for five features), resulting in a small memory footprint and transparent behavior. While strict normality of RMS-derived features was not explicitly verified, empirical results indicate sufficient separation between normal and anomalous behavior for this application. Evaluation was limited to a single appliance with controlled anomalies; broader validation remains future work.

## IX. CONCLUSION & FUTURE WORK

This paper presented a fully autonomous Z-Score–based anomaly detection system implemented on a resource-constrained STM32 microcontroller. The system performs on-device training and inference using RMS current side-channel measurements to characterize appliance behavior and detect anomalous events. A 14-day real-world deployment on a household refrigerator demonstrated accurate detection, low inference latency, and minimal memory overhead.

Future work will investigate complementary lightweight detection techniques, adaptive statistical modeling to handle evolving operating conditions, and evaluation across additional appliances and deployment environments, while preserving the MCU-resident design demonstrated in this work.

ACKNOWLEDGMENT

The authors would like to thank the Saudi Arabian Cultural Mission (SACM) for supporting the author's academic program, and the SMART Lab at Wright State University for providing guidance and technical resources.